\documentclass[sigconf]{acmart}
\AtBeginDocument{%
  }

\renewcommand\footnotetextcopyrightpermission[1]{}
\setcopyright{none}
\copyrightyear{}
\acmYear{}
\acmDOI{}
\acmConference{}{}{}
\acmISBN{}

\usepackage{multirow}
\usepackage{epsfig}
\usepackage{amsmath}
\usepackage{amsfonts}
\usepackage{subfigure}
\usepackage{marvosym}
\usepackage{color}
\usepackage{threeparttable}
\usepackage{algorithm}
\usepackage{algpseudocode}
\usepackage{amsthm}
\usepackage{pifont}
\usepackage{transparent}
\usepackage{makecell}

\begin{document}

\title{Aura: Universal Multi-dimensional Exogenous Integration \\
for Aviation Time Series}

\author{Jiafeng Lin}
\authornote{Both authors contributed equally to this research.}
\email{lin-jf21@mails.tsinghua.edu.cn}
\orcid{0009-0004-5299-9992}
\affiliation{%
  \institution{School of Software, BNRist, \\ Tsinghua University}
  \city{Beijing}
  \country{China}
}
\author{Mengren Zheng}
\authornotemark[1]
\email{mengrenzheng@gmail.com}
\orcid{0009-0002-0984-208X}
\affiliation{%
  \institution{Hongshen College, \\ Chongqing University}
  \city{Chongqing}
  \country{China}
}
\author{Simeng Ye}
\email{ysm23@mails.tsinghua.edu.cn}
\orcid{0009-0008-4773-8332}
\affiliation{%
  \institution{School of Software, \\ Tsinghua University}
  \city{Beijing}
  \country{China}
}
\author{Yuxuan Wang}
\email{wangyuxu22@mails.tsinghua.edu.cn}
\orcid{0000-0002-4899-4716}
\affiliation{%
  \institution{School of Software, BNRist, \\ Tsinghua University}
  \city{Beijing}
  \country{China}
}
\author{Huan Zhang}
\email{zhanghuan_a@csair.com}
\affiliation{%
  \institution{China Southern Airlines}
  \city{Guangzhou}
  \country{China}
}
\author{Yuhui Liu}
\email{liuyh@csair.com}
\affiliation{%
  \institution{China Southern Airlines}
  \city{Guangzhou}
  \country{China}
}
\author{Zhongyi Pei}
\authornote{Corresponding Author.}
\email{peizhyi@tsinghua.edu.cn}
\affiliation{%
  \institution{School of Software, BNRist, \\ Tsinghua University}
  \city{Beijing}
  \country{China}
}
\author{Jianmin Wang}
\email{jimwang@tsinghua.edu.cn}
\affiliation{%
  \institution{School of Software, BNRist, \\ Tsinghua University}
  \city{Beijing}
  \country{China}
}

\renewcommand{\shortauthors}{Lin et al.}

\begin{abstract}
Time series forecasting has witnessed an increasing demand across diverse industrial applications, where accurate predictions are pivotal for informed decision-making.  Beyond numerical time series data, reliable forecasting in practical scenarios requires integrating diverse exogenous factors. Such exogenous information is often multi-dimensional or even multimodal, introducing heterogeneous interactions that unimodal time series models struggle to capture. In this paper, we delve into an aviation maintenance scenario and identify three distinct types of exogenous factors that influence temporal dynamics through distinct interaction modes. Based on this empirical insight, we propose \textbf{Aura}, a universal framework that explicitly organizes and encodes heterogeneous external information according to its interaction mode with the target time series. Specifically, Aura utilizes a tailored tripartite encoding mechanism to embed heterogeneous features into well-established time series models, ensuring seamless integration of non-sequential context. Extensive experiments on a large-scale, three-year industrial dataset from China Southern Airlines, covering the Boeing 777 and Airbus A320 fleets, demonstrate that Aura consistently achieves state-of-the-art performance across all baselines and exhibits superior adaptability. Our findings highlight Aura's potential as a general-purpose enhancement for aviation safety and reliability. \footnote{Preliminary work.}
\end{abstract}



\maketitle

\section{Introduction}

\begin{figure*}[h]
  \centering
  \includegraphics[width=\linewidth]{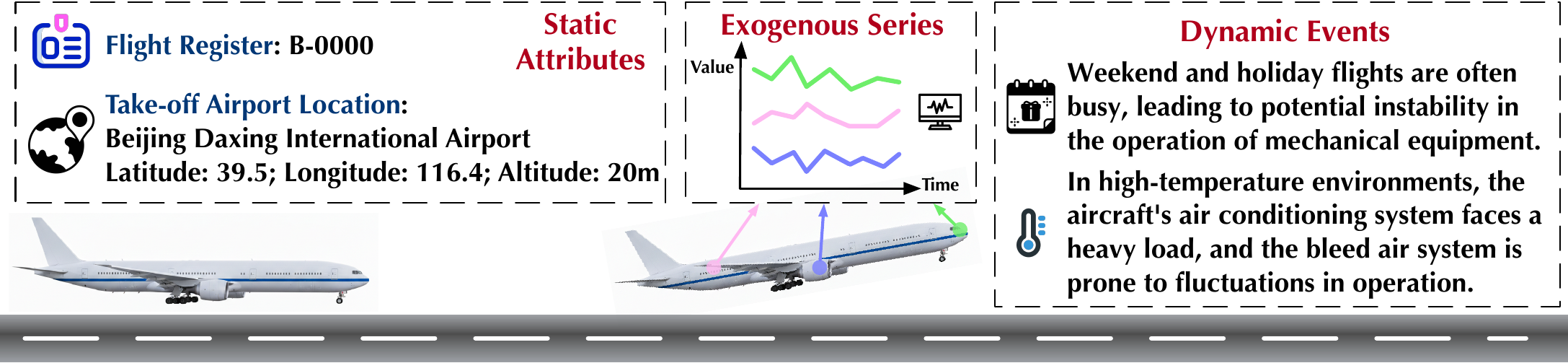}
  \caption{Illustration of diverse exogenous information influencing aircraft component anomaly detection. Beyond the time series directly associated with the equipment, this detection process incorporates multiple factors, including correlated sensor signals, aircraft register, airport location, and dynamic environmental takeoff conditions, which collectively influence the temporal dynamics, necessitating a unified integration for accurate health monitoring.}
  \label{fig:intro}
\end{figure*}

Time series forecasting \cite{brockwell2002introduction, de200625} plays a pivotal role in supporting informed decision making across a wide range of real-world applications, including finance \cite{sezer2020financial}, energy \cite{deb2017review}, and transportation \cite{karami2020smart}. In recent years, rapid advances in deep learning have driven a shift from classical statistical methods to deep time series models, which demonstrate superior capacity to capture complex temporal dynamics and long-range dependencies \cite{zhou2021informer, wang2024deep}. In particular, Transformer has emerged as a dominant architecture due to its strong sequence modeling capability and scalability \cite{wu2021autoformer, nie2022time, liu2023itransformer, zhang2023crossformer}. Despite these successes, most existing approaches rely primarily on historical observations of the target series and largely overlook rich exogenous information available in practical scenarios.

In real-world scenarios, temporal dynamics are rarely governed solely by historical patterns. Similar temporal trends may convey fundamentally different meanings depending on domain-specific context \cite{dong2024metadata, williams2024context}. For example, an increasing trend may indicate healthy economic growth, an emerging equipment fault, or abnormal demand surges in energy systems. Models that rely solely on historical observations struggle to disambiguate such scenarios, potentially resulting in inaccurate forecasts or even risky decisions. This limitation underscores the need to incorporate exogenous information beyond raw time series data.

Notably, exogenous information in real-world forecasting scenarios is heterogeneous not only in modality and semantics, but also in its interaction with the target time series data. Such information may include static metadata describing intrinsic properties of the system, dynamic contextual events expressed as unstructured text, and exogenous series that align and co-evolve with the target series. As shown in Figure~\ref{fig:intro}, in aviation forecasting, these factors range from static aircraft specifications and configuration parameters to dynamic textual reports reflecting weather conditions or operational events, as well as multiple correlated sensor measurements. Each type of exogenous factor influences temporal dynamics through a distinct interaction mode, making unified integration particularly challenging.

Existing approaches for exogenous time series forecasting have predominantly focused on incorporating exogenous time series data \cite{wang2024timexer, qiu2025dag, zhang2026ditsmultimodaldiffusiontransformers}. While effective for modeling inter-series dependencies, these methods support only numerical inputs and are not designed to accommodate unstructured or irregularly occurring exogenous information, such as text. In practice, textual signals are often misaligned with the target series and exhibit indirect and semantically implicit relationships with temporal dynamics, which are not well supported by existing forecasting frameworks.

Motivated by this observation, we study the problem of integrating multi-dimensional exogenous information into deep time series forecasting models. Focusing on a practical aviation predictive maintenance scenario, we propose Aura, a Transformer-based framework that explicitly models different types of exogenous factors according to their roles in time series forecasting. Specifically, Aura integrates heterogeneous external information at different stages of the Transformer architecture, enabling effective fusion without substantial architectural modifications. We evaluate Aura on a large-scale, three-year industrial dataset collected from China Southern Airlines, covering both Boeing 777 and Airbus A320 fleets. Extensive experimental results demonstrate that Aura consistently outperforms a wide range of strong baselines and exhibits robust generalization across various aircraft types and forecasting tasks. These results underscore the effectiveness of Aura as a practical framework for improving time series forecasting in real-world aviation predictive maintenance settings. Our contributions are summarized as follows:
\begin{itemize}
    \item We identify and empirically validate three types of exogenous factors that play different roles in time series forecasting, each exhibiting a distinct interaction with the target series, as demonstrated in an aviation predictive maintenance setting.
    \item We propose Aura, a universal framework that introduces specialized encoding designs for three types of exogenous factors, enabling targeted integration according to their distinct roles in time series forecasting.
    \item Extensive experiments on a large-scale, real-world industrial aviation dataset show that Aura consistently outperforms state-of-the-art deep time series models across diverse forecasting scenarios.
\end{itemize}

\section{Related Work}

\subsection{Time Series Forecasting With Exogenous Variables}

Time series forecasting with exogenous variables has been extensively studied for decades. Early efforts primarily focused on extending classical statistical models to accommodate exogenous variables. Representative examples include ARIMAX \cite{williams2001multivariate} and SARIMAX \cite{mulla2024times}, which incorporate exogenous variables as additive components within linear stochastic frameworks. While effective in certain scenarios, these approaches rely on strong parametric assumptions and linear constraints, which significantly limit their ability to model the complex, nonlinear interactions between endogenous dynamics and external drivers commonly observed in real-world data.

With the advent of deep learning, a growing body of research has explored neural approaches for integrating exogenous time series. Early approaches such as TiDE \cite{das2023long} and NBEATSx \cite{olivares2023neural} adopt a holistic fusion strategy, directly concatenating the target and covariate series and mapping them to high-dimensional representations. Although conceptually simple, such black-box fusion mechanisms often struggle to disentangle informative signals from irrelevant or noisy covariates. To address this limitation, Temporal Fusion Transformer (TFT) \cite{lim2021temporal} introduces variable selection networks and attention mechanisms to dynamically modulate covariate importance, while TimeXer \cite{wang2024timexer} leverages cross-attention to explicitly model interactions between target and exogenous series. More recently, CrossLinear \cite{zhou2025crosslinear} mitigates overfitting by employing a lightweight cross-correlation module to capture stable and interpretable variable relationships. DUET \cite{qiu2025duet} further advances this line of research by jointly addressing temporal heterogeneity and complex inter-channel interactions through a dual-clustering framework, combining frequency-domain metric learning with sparsification to refine the coupling between endogenous and exogenous variables.

\subsection{Multimodal Time Series Forecasting}

Driven by the rapid progress of multimodal foundation models, multimodal time series forecasting has recently attracted increasing attention \cite{jiang2025multi}.  Within this line of research, Transformer-based architectures \cite{vaswani2017attention} have emerged as the \textit{de facto} architectural choice, owing to their strong sequence modeling capability and their central role in large language models. Along this direction,
GPT4TS \cite{zhou2023one} establishes a foundational approach by utilizing frozen pre-trained LLMs as universal backbones, demonstrating that language models can be effectively adapted to time series analysis with minimal fine-tuning. Building on this, Time-LLM \cite{jin2023time} introduces a reprogramming framework that employs text prototypes and prompts to align time series patches with language embeddings while preserving the backbone intact.  Similarly, $\text{S}^2$IP-LLM \cite{pan2024s} aligns the pre-trained semantic space with the time series embedding space and performs time series forecasting based on learned prompts from the joint space. To further address domain heterogeneity, UniTime \cite{liu2024unitime} enables cross-domain generalization through instruction tuning and a Language Time Series Transformer, mitigating discrepancies in data distributions across diverse applications. More recently, TimeCMA \cite{liu2025timecma} introduces a dual-branch architecture that achieves fine-grained modality alignment by disentangling temporal representations and prompt embeddings, while improving efficiency by compressing essential multimodal information into a single token.

Given the critical role of textual data in conveying information about external events that can significantly influence time series dynamics \cite{williams2024context, chang2025time, wang2025chattime}, several text-enhanced forecasting approaches have been proposed. From News to Forecast \cite{wang2024news} leverages LLM-based agents to adaptively integrate social and news events into forecasting models, aligning textual content with temporal fluctuations to yield more informative predictions. Time-MMD \cite{liu2024time} introduces a model-agnostic multimodal framework that independently models numerical time series and textual modalities, and combines them via a learnable linear weighting mechanism to yield the final prediction.

\section{Data Mining}

In this section, we present the data-mining strategy for identifying multi-dimensional factors corresponding to the operational health of the Pressure Regulating and Shut-Off Valve (PRSOV). We systematically integrate endogenous physical time series data with exogenous operational and environmental contexts to construct a comprehensive representation of valve health under real-world operating conditions.

\begin{figure}[h]
  \centering
  \includegraphics[width=0.95\linewidth]{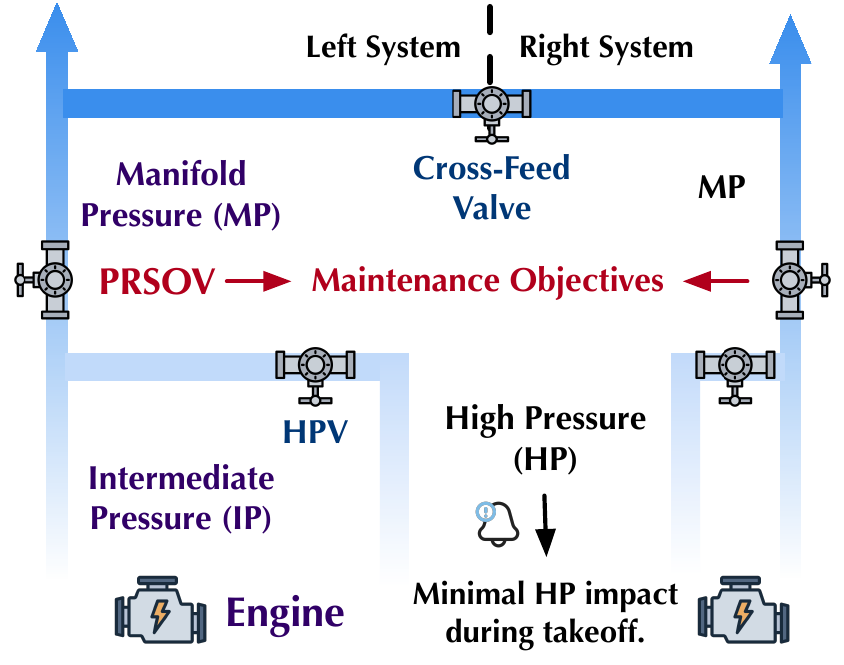}
  \caption{Bleed Air System Schematic.}
  \label{prsov}
  \vspace{-10pt}
\end{figure}

\paragraph{Flight Phase Extraction}

To ensure the fidelity of the maintenance forecasting model, we perform a targeted extraction of the \emph{take-off phase} as the primary observation window. This phase corresponds to the operating regime with the highest pneumatic load, during which the bleed air system operates under peak engine thrust. Under such conditions, deviations in pressure regulation are more likely to reflect intrinsic valve degradation rather than benign operational variability. By restricting the analysis to this phase, we obtain a consistent and stress-sensitive view of PRSOV behavior.

\paragraph{Domain-Driven Variates Selection}

The goal of the mining process is to establish a robust mapping between the mechanical health of the valve and its observable system behavior. We select Manifold Pressure (MP) as the target endogenous variable. Within the bleed air system architecture, the fundamental function of PRSOV is to regulate and modulate manifold pressure. Accordingly, MP serves as a direct physical proxy for valve performance based on aeronautical domain knowledge. To account for external influences and explicitly condition manifold pressure on engine operating states, we further include Engine Speed ($\text{N}_2$) and Intermediate-Stage Pressure (IP) as the exogenous series. These variables reflect the boundary conditions of the bleed air system, enabling the model to disentangle pressure variations caused by engine state changes from those arising from intrinsic valve instability.

\paragraph{Contextual Mining}
Beyond intrinsic system measurements, the operational state of the PRSOV is influenced by a range of external contextual factors.
Based on this consideration, we incorporate additional contextual information to provide a comprehensive view of the health state of the equipment. Aircraft Registration is used as an equipment identifier to capture asset-specific characteristics. Operational and environmental contexts are constructed with timestamps, departure airports, and GPS coordinates, enabling the integration of ambient conditions such as temperature and humidity.
These environmental factors influence the load of the aircraft's air conditioning system and, consequently, the demand on the bleed air subsystem. In addition, calendar-based information, including holidays and peak travel periods, is incorporated to characterize high-intensity operational cycles associated with increased component stress. Collectively, these contextual features provide complementary evidence for explaining the performance variability of the PRSOV under complex real-world operating conditions.

\begin{figure*}[h]
  \centering
  \includegraphics[width=\linewidth]{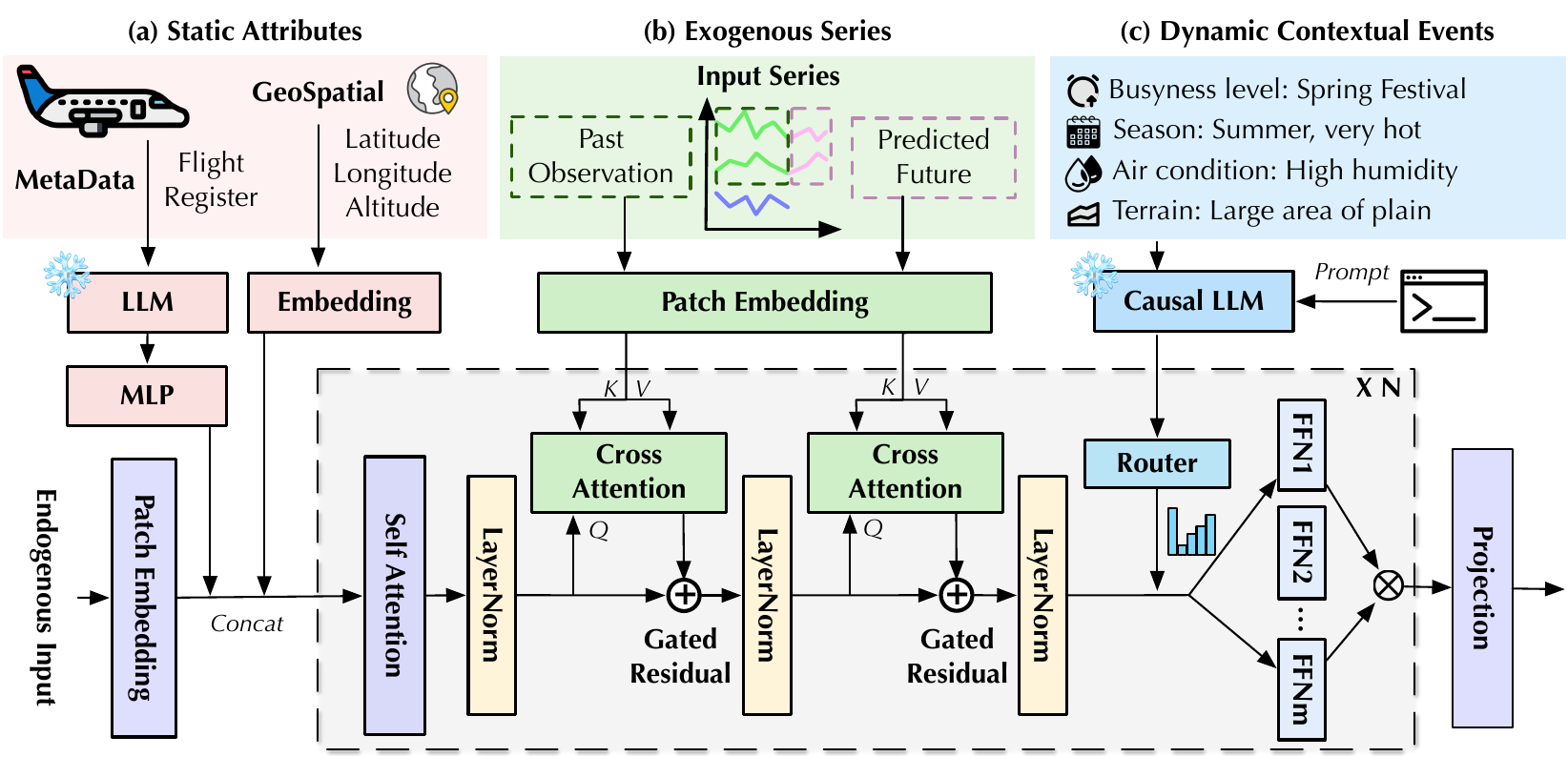}
  \caption{Overall architecture of Aura, a multi-dimensional exogenous integration framework. Aura integrates static attributes encoded by LLMs and geospatial embeddings, and prepends them to endogenous tokens. Two cross-attention layers fuse exogenous series via residual connections to regulate the strength of integration. A Mixture of Experts (MoE) module leverages LLM-generated future insights from dynamic events to guide time series forecasting.}
  \label{fig:pipeline}
\end{figure*}

\section{Problem Formulation}

We study an exogenous-informed time series forecasting problem in which a univariate endogenous time series is influenced by complex physical mechanisms and multi-dimensional external contexts. Given the historical observations of the endogenous series $\mathbf{x}_{1:T} = \{x_1, x_2, \dots, x_T\} \in \mathbb{R}^{T \times 1}$, our objective is to leverage three distinct types of external information:

\begin{enumerate}
    \item \emph{Static Attributes}: This category, denoted as $\mathcal{S}$, includes time-invariant attributes associated with the entity, such as the aircraft register and geographic coordinates. These attributes are encoded as static embeddings that provide identity- and location-aware grounding for the model.
    \item \emph{Exogenous Series}: These consist of multiple exogenous series $\mathbf{Z} = \{\mathbf{Z}_{1:T}^{his}, \mathbf{Z}_{T+1:T+S}^{fut}\} \in \mathbb{R}^{(T+S) \times D}$, where $D$ is the number of variables. Given that exogenous conditions shape the trajectory of the endogenous variable, these series serve as essential conditioning inputs. They allow the model to capture the intrinsic physical dependencies and law-like correlations that govern how the system evolves over time.
    \item \emph{Dynamic Contextual Events}: To capture the evolving operational environment, we incorporate textual information $\mathcal{T}$ including ambient conditions, maintenance logs, or external event descriptions.
    Unlike numerical time series data, $\mathcal{T}$ provides semantic insights that modulate the underlying temporal dynamics.
\end{enumerate}

The goal is to learn a forecasting model $\mathcal{F}_\theta$ that predicts the future values of the endogenous series $\hat{\mathbf{x}}_{T+1:T+S} \in \mathbb{R}^{S \times 1}$ over the forecasting horizon $S$. Formally, the task is defined as:
\begin{equation}
    \hat{\mathbf{x}}_{T+1:T+S} = \mathcal{F}_\theta \left( \mathbf{x}_{1:T}, \mathbf{Z}_{1:T}^{his}, \mathbf{Z}_{T+1:T+S}^{fut}, \mathcal{T}, \mathcal{S} \right)
\end{equation}

\section{Aura}
  
\subsection{Overview}

As illustrated in Figure~\ref{fig:pipeline}, we propose Aura, a unified time series forecasting framework that explicitly incorporates multi-dimensional exogenous information, including static attributes, exogenous series, and external events. Technologically, Aura focuses on forecasting the endogenous variable, while static attributes, exogenous series, and external events are leveraged as auxiliary factors that provide additional contextual information. To accommodate the heterogeneous nature and varying influence of exogenous data, Aura employs a hierarchical integration strategy that assigns each type of external information a dedicated fusion mechanism. Concretely,  static attributes, exogenous series, and dynamic event information are integrated through modality-specific modules, including direct feature fusion, cross-attention mechanisms, and a Mixture-of-Experts (MoE) module. This design enables the model to selectively emphasize relevant contextual information and adaptively modulate its contributions to the forecasting process.

\subsection{Time Series Embedding}
We follow the convention \cite{nie2022time} and employ non-overlapping patch embedding for \emph{both} endogenous and exogenous time series. By adopting a shared patch granularity across all temporal inputs, the model avoids scale and alignment mismatches between the endogenous and exogenous series and establishes a unified patch-level representation for subsequent self-attention and cross-attention fusion. Given an input sequence $\mathbf{x}\in $ $\mathbb{R}^{T}$, we divide it into $N$ non-overlapping patches of length $P$. If $T$ is not divisible by $P$, zero-padding is applied to the end of the sequence, resulting in $N=\lceil T/P\rceil$ patches. Each patch is then projected into a $D$-dimensional temporal token through a shared trainable projector $\mathrm{PatchEmbed}(\cdot):\mathbb{R}^{P}\rightarrow\mathbb{R}^{D}$:
\begin{align}
&\{\mathbf{s}_1,\mathbf{s}_2,\ldots,\mathbf{s}_N\}=\mathrm{Patchify}(\mathbf{x}),\qquad \mathbf{s}_i\in\mathbb{R}^{P}, \\
&\{\mathbf{h}_i\}_{i=1}^{N}=\mathrm{PatchEmbed}\big(\{\mathbf{s}_i\}_{i=1}^{N}\big),\qquad 
\mathbf{h}_i=\mathbf{W}\mathbf{s}_i+\mathrm{PE}(i),
\end{align}
where $\mathbf{W}\in\mathbb{R}^{D\times P}$ is shared across patches and $\mathrm{PE}(i)$ denotes the positional embedding of the $i$-th patch. For multivariate exogenous inputs, the patch embedding is applied independently to each variable, and the resulting exogenous tokens are concatenated.

\subsection{Static Attributes Embedding}

To incorporate instance-level static information beyond temporal observations, we encode metadata as a small set of \emph{meta tokens}, including the aircraft identifier and the takeoff geolocation (latitude, longitude, altitude). These meta tokens, which share the same hidden dimension as the time series tokens, are concatenated with the endogenous patch tokens, enabling joint modeling via self-attention and cross-attention mechanisms.

\paragraph{Meta Embedding}
Given the Aircraft Registration text $\mathcal{S}$, we obtain token-level representations using a pretrained language model, which are then aggregated via average pooling to form a fixed-length vector. This vector is then projected onto the hidden dimension of the time series patch tokens for alignment. The formalization is given by:
\begin{equation}
\bar{\mathbf{H}}_{a}=\mathrm{AvgPool}\big(\mathrm{LLM}(\mathcal{S})\big),\qquad 
\mathbf{z}_{a}=\mathrm{Proj}_{a}\big(\bar{\mathbf{H}}_{a}\big)\in\mathbb{R}^{D}.
\end{equation}
Here, $\mathrm{LLM}(\cdot)$ denotes a frozen BERT \cite{devlin2019bert} model,  $\bar{\mathbf{H}}_{a}$ represents the average pooling output, and $\mathrm{Proj}_{a}(\cdot)$ is the linear projection that maps it into the shared embedding space of hidden dimension $D$.

\paragraph{GeoSpatial Embedding}
The takeoff location is represented as a three-dimensional vector $\mathbf{g}=(\mathrm{lat}, \mathrm{lon}, \mathrm{alt})\in\mathbb{R}^{3}$ corresponding to the latitude, longitude, and altitude. We project this vector into the hidden dimension of $D$, ensuring a consistent token interface across heterogeneous inputs:
\begin{equation}
\mathbf{z}_{g}=\mathrm{Proj}_{g}(\mathbf{g})\in\mathbb{R}^{D}.
\end{equation}
Here, $\mathrm{Proj}_{g}(\cdot)$ denotes the learnable projection for geospatial data.

\paragraph{Meta Token Concatenation}
After embedding the aircraft identifier and geospatial attributes, the resulting meta tokens are appended to the historical time-series patch tokens in a fixed order. This yields the final input sequence $\mathbf{H}_{\mathrm{endo}}$:
\begin{equation}
\mathbf{H}_{\mathrm{endo}}=\big[\mathbf{h}_{\mathrm{endo}};\mathbf{z}_{a};\mathbf{z}_{g}\big]\in\mathbb{R}^{(N+2)\times D},
\end{equation}
where $\mathbf{h}_{\mathrm{endo}}\in\mathbb{R}^{N\times D}$ denotes the historical endogenous patch tokens, $\mathbf{z}_{a}$ denotes the embedded token of static metadata, $\mathbf{z}_{g}$ denotes the embedded geospatial token. The concatenated sequence is then fed into subsequent Transformer layers, allowing static instance-level information to interact directly with temporal dynamics.

\subsection{Self Attention}

Given the concatenated input tokens $\mathbf{H}_{\mathrm{endo}}\in\mathbb{R}^{(N+2)\times D}$, we employ self-attention to effectively capture temporal dependencies among the endogenous tokens. This approach enables each token to dynamically aggregate information from all others, capturing both short-term fluctuations and long-range relationships. Additionally, tokens enriched with static attributes offer comprehensive global context to the endogenous tokens.
\begin{equation}
\mathrm{Attention}(\mathbf{H})
=
\mathrm{Softmax}\!\left(\frac{\mathrm{Mask}(\mathbf{Q}\mathbf{K}^{\top})}{\sqrt{d}}\right)\mathbf{V},
\end{equation}
Here $\mathbf{Q}$, $\mathbf{K}$, and $\mathbf{V}$ are obtained via linear projections of $\mathbf{H}$, $d$ denotes the hidden dimension of each attention head, and $\mathrm{Mask}(\cdot)$ represents an optional attention mask.

\begin{table*}[htbp]
  \caption{Results of time series forecasting and anomaly detection on China Southern Airlines datasets. \textbf{Bold} indicates the best performance. True Alert Rates (TAR) are calculated under a constrained false alarm rate of 5\% to ensure practical reliability.}
  \vspace{-5pt}
  \renewcommand{\arraystretch}{1}
  \centering
  \begin{threeparttable}
  \renewcommand{\multirowsetup}{\centering}
  \setlength{\tabcolsep}{2.2pt}
  \begin{tabular}{c | c | cccccccccccccc}
    \toprule

    \multicolumn{2}{c}{\multirow{2}{*}{\textbf{Model}}} & 
    \scalebox{0.85}{\textbf{Aura}} & 
    \scalebox{0.8}{DUET} &           
    \scalebox{0.8}{CrossLinear} &    
    \scalebox{0.8}{Timer-XL} &     
    \scalebox{0.8}{TimeXer} &        
    \scalebox{0.8}{TiDE} &           
    \scalebox{0.8}{TimeLLM} &    
    \scalebox{0.8}{GPT4TS} &    
    \scalebox{0.8}{iTransformer} &
    \scalebox{0.8}{DLinear} &     
    \scalebox{0.8}{PatchTST} &    
    \scalebox{0.8}{TimesNet} &    
    \scalebox{0.8}{Autoformer} &  
    \scalebox{0.8}{LightGBM} \\   

    \multicolumn{2}{c}{} & 
    \scalebox{0.8}{\textbf{(Ours)}} &                  
    \scalebox{0.8}{(\citeyear{qiu2025duet})} &         
    \scalebox{0.8}{(\citeyear{zhou2025crosslinear})} & 
    \scalebox{0.8}{(\citeyear{liu2024timer})} &  
    \scalebox{0.8}{(\citeyear{wang2024timexer})} &  
    \scalebox{0.8}{(\citeyear{das2023long})} &         
    \scalebox{0.8}{(\citeyear{jin2023time})} &   
    \scalebox{0.8}{(\citeyear{zhou2023one})} &   
    \scalebox{0.8}{(\citeyear{liu2023itransformer})} &  
    \scalebox{0.8}{(\citeyear{zeng2023transformers})} & 
    \scalebox{0.8}{(\citeyear{nie2022time})} & 
    \scalebox{0.8}{(\citeyear{wu2022timesnet})} &
    \scalebox{0.8}{(\citeyear{wu2021autoformer})} &
    \scalebox{0.8}{(\citeyear{ke2017lightgbm})} \\
    \toprule 

    Boeing & MSE $\downarrow$ & \textcolor{red}{\textbf{0.075}} & 0.157 & 0.109 & 0.140 & 0.093 & 0.318 & 0.358 & 0.158 & \textcolor{blue}{\underline{0.086}} & 0.203 & 0.122 & 0.113 & 0.823 & 0.145 \\
    \cmidrule{2-16}
    777 & MAE $\downarrow$ & \textcolor{red}{\textbf{0.180}} & 0.250 & 0.225 & 0.232 & 0.197 & 0.401 & 0.427 & 0.240 & \textcolor{blue}{\underline{0.184}} & 0.286 & 0.224 & 0.208 & 0.711 & 0.220 \\
    \cmidrule{2-16}
    Left & TAR $\uparrow$ & \textcolor{red}{\textbf{0.625}} & \textcolor{blue}{\underline{0.500}} & \textcolor{blue}{\underline{0.500}} & \textcolor{blue}{\underline{0.500}} & \textcolor{blue}{\underline{0.500}} & 0.375 & 0.375 & \textcolor{blue}{\underline{0.500}} & \textcolor{blue}{\underline{0.500}} & \textcolor{blue}{\underline{0.500}} & \textcolor{blue}{\underline{0.500}} & \textcolor{blue}{\underline{0.500}} & 0.375 & \textcolor{blue}{\underline{0.500}} \\
    \midrule

    Boeing & MSE $\downarrow$ & \textcolor{red}{\textbf{0.086}} & 0.103 & 0.117 & 0.118 & 0.101 & 0.308 & 0.413 & 0.100 & \textcolor{blue}{\underline{0.090}} & 0.170 & 0.107 & 0.106 & 0.668 & 0.096 \\
    \cmidrule{2-16}
    777 & MAE $\downarrow$ & \textcolor{red}{\textbf{0.182}} & 0.202 & 0.230 & 0.213 & 0.201 & 0.396 & 0.466 & \textcolor{blue}{\underline{0.189}} & \textcolor{blue}{\underline{0.189}} & 0.262 & 0.207 & 0.199 & 0.636 &  \textcolor{red}{\textbf{0.182}} \\
    \cmidrule{2-16}
    Right & TAR $\uparrow$ & \textcolor{red}{\textbf{0.625}} & \textcolor{blue}{\underline{0.500}} & 0.375 & 0.375 & 0.375 & \textcolor{red}{\textbf{0.625}} & \textcolor{red}{\textbf{0.625}} & \textcolor{blue}{\underline{0.500}} & \textcolor{red}{\textbf{0.625}} & \textcolor{red}{\textbf{0.625}} & \textcolor{red}{\textbf{0.625}} & 0.375 & \textcolor{red}{\textbf{0.625}} & \textcolor{red}{\textbf{0.625}} \\
    \midrule

    Airbus & MSE $\downarrow$ & \textcolor{red}{\textbf{0.048}} & 0.058 & 0.065 & \textcolor{blue}{\underline{0.053}} & 0.054 & 0.098 & 0.107 & 0.059 & 0.056 & 0.094 & 0.065 & 0.059 & 0.174 & 0.061 \\
    \cmidrule{2-16}
    A320 & MAE $\downarrow$ & \textcolor{red}{\textbf{0.138}} & 0.148 & 0.147 & 0.144 & 0.143 & 0.182 & 0.202 & 0.147 & \textcolor{blue}{\underline{0.142}} & 0.179 & 0.172 & 0.151 & 0.263 & 0.142 \\
    \cmidrule{2-16}
    Left & TAR $\uparrow$ & \textcolor{red}{\textbf{0.345}} & 0.276 & 0.172 & \textcolor{blue}{\underline{0.310}} & 0.276 & 0.241 & 0.276 & \textcolor{blue}{\underline{0.310}} & 0.276 & 0.241 & \textcolor{blue}{\underline{0.310}} & \textcolor{blue}{\underline{0.310}} & 0.241 & \textcolor{blue}{\underline{0.310}} \\
    \midrule

    Airbus & MSE $\downarrow$ & \textcolor{red}{\textbf{0.063}} & 0.072 & 0.069 & 0.069 & \textcolor{blue}{\underline{0.066}} & 0.105 & 0.106 & 0.069 & 0.067 & 0.105 & 0.084 & 0.069 & 0.320 & 0.070 \\
    \cmidrule{2-16}
    A320 & MAE $\downarrow$ & \textcolor{red}{\textbf{0.146}} & 0.152 & 0.150 & 0.152 & 0.150 & 0.189 & 0.207 & 0.152 & \textcolor{blue}{\underline{0.148}} & 0.186 & 0.196 & 0.151 & 0.411 & 0.149 \\
    \cmidrule{2-16}
    Right & TAR $\uparrow$ & \textcolor{red}{\textbf{0.414}} & 0.310 & 0.276 & 0.310 & 0.310 & 0.310 & \textcolor{blue}{\underline{0.379}} & \textcolor{blue}{\underline{0.379}} & 0.310 & 0.310 & 0.276 & 0.310 & 0.276 & 0.310 \\

    \bottomrule
  \end{tabular}
  \end{threeparttable}
\label{tab:main_result}
\end{table*}

\subsection{Exogenous Cross Attention}

Aura further incorporates multi-dimensional exogenous information using \emph{cross-attention}. Specifically, we employ two distinct cross-attention blocks to integrate both historical and predicted-future exogenous series. These exogenous series are embedded as exogenous patch tokens, which serve as key-value pairs to interact with the endogenous tokens.

Let $\mathbf{H}_{\mathrm{endo}}\in\mathbb{R}^{N\times D}$ denote the endogenous tokens after self-attention, and $\mathbf{h}^h_{\mathrm{exog}}$ and $\mathbf{h}^f_{\mathrm{exog}}$ be the embedded historical and future exogenous tokens, respectively. We perform two stages of cross-attention to incorporate historical and future exogenous series:
\begin{equation}
\begin{aligned}
\mathbf{H}'_{\mathrm{endo}}=\mathrm{Attn}\!\left(\mathbf{H}_{\mathrm{endo}},\,\mathbf{h}^h_{\mathrm{exog}},\,\mathbf{h}^h_{\mathrm{exog}}\right),\\
\mathbf{H}''_{\mathrm{endo}}=\mathrm{Attn}\!\left(\mathbf{H}'_{\mathrm{endo}},\,\mathbf{h}^f_{\mathrm{exog}},\,\mathbf{h}^f_{\mathrm{exog}}\right),
\end{aligned}
\end{equation}
where $\mathrm{Attn}$ denotes the cross-attention between endogenous and exogenous tokens.

To stabilize exogenous fusion, we introduce a lightweight gated residual that adaptively modulates the contribution of each cross-attention branch. Specifically, the output of each cross-attention block is scaled by a sample-wise coefficient $\alpha\in(0,1)$, which is computed from a learnable function of the query. This coefficient $\alpha$ is then applied to the cross-attention output before being added back via a residual connection. This gating mechanism enables selective integration of informative exogenous series while mitigating over-conditioning from noisy or weakly relevant information. 

\subsection{External Mixture of Experts}

In real-world industrial applications, the production environment evolves over time, and dynamic environmental events influence the time series dynamics. The causal reasoning capabilities of LLMs can interpret environmental events and infer their effects on time series data, thereby enabling more accurate predictions. Motivated by this capability, we leverage an LLM to generate inferences about future time series dynamics and incorporate the resulting representations into a Mixture of Experts (MoE) framework. Specifically, the inferred future insight is fed into a gating network that guides expert selection for time series forecasting. The gating network sparsely selects a subset of experts that are shared across all endogenous tokens. This design integrates future-relevant external knowledge directly into the forecasting pipeline, without explicit modality fusion, yielding more informative predictions.

Formally, given the dynamic text sequence $\mathcal{T}$, we first process it through a pre-trained LLM with a meticulously designed prompt to obtain hidden representations that encode inferred future temporal variations. The hidden representations are aggregated via average pooling and passed to a gating network that applies a softmax to assign weights to each expert. The weighted combination of experts is then used to refine the time series prediction. The process is formalized as follows:
\begin{align}
\bar{\mathbf{H}} = \text{AvgPool}\left(\mathrm{LLM}(\mathcal{T})\right)&, \quad
\mathbf{w} = \text{Softmax}\left(\mathbf{W}_g \cdot \bar{\mathbf{H}} + \mathbf{b}_g\right), \\
\mathrm{MoE}(\mathbf{H}_{\mathrm{endo}}) &= \sum_{i=1}^{K} w_i \cdot \text{FFN}_i(\mathbf{H}_{\mathrm{endo}}),
\end{align}
where $\mathcal{T}$ is the dynamic textual input, $\bar{\mathbf{H}}$ is the pooled hidden representation from the LLM, $\mathbf{W}_g$ and $\mathbf{b}_g$ are the learnable parameters of the gating network, $\mathbf{w}$ is the gating weight, $K$ is the number of experts, $\text{FFN}_i$ denotes the $i$-th expert's feed-forward network, and $\mathbf{H}_{\mathrm{endo}}$ is the endogenous tokens. The output of the MoE block is a weighted sum of the experts, allowing the model to dynamically select which experts to use based on the given textual input.

\section{Experiment}

\paragraph{Datasets} The main experiments are conducted on a large-scale, three-year dataset collected from China Southern Airlines, covering 99 aircraft from Boeing 777 and Airbus A320 fleets. We specifically extract segments corresponding to the takeoff phase and perform independent analyses for the left and right systems. Additionally, the publicly available Electricity Price Forecasting (EPF) benchmark \cite{lago2021forecasting} is included to evaluate Aura's performance under the forecasting with exogenous variables.

\begin{table*}[t]
\setlength{\tabcolsep}{2.4pt}
\renewcommand{\arraystretch}{1}
\caption{Exogenous forecasting results on the EPF benchmark with future exogenous series. We set the input length to 168 and the prediction length to 24, following TimeXer \cite{wang2024timexer} experimental setup. \emph{AVG} means the average results from all five datasets.}

\begin{scriptsize}
\begin{tabular}{c|cc|cc|cc|cc|cc|cc|cc|cc|cc|cc|cc} 
\toprule
\scalebox{1.2}{Model}   & \multicolumn{2}{c}{\scalebox{1.2}{\textbf{Aura}}} & \multicolumn{2}{c}{\scalebox{1.2}{iTransformer}} & \multicolumn{2}{c}{\scalebox{1.2}{TimeXer}} & \multicolumn{2}{c}{\scalebox{1.2}{TimerXL}} & \multicolumn{2}{c}{\scalebox{1.2}{TiDE}} & \multicolumn{2}{c}{\scalebox{1.2}{CrossLinear}} & \multicolumn{2}{c}{\scalebox{1.2}{DUET}} & \multicolumn{2}{c}{\scalebox{1.2}{PatchTST}} & \multicolumn{2}{c}{\scalebox{1.2}{TimesNet}} & \multicolumn{2}{c}{\scalebox{1.2}{DLinear}} & \multicolumn{2}{c}{\scalebox{1.2}{Autoformer}} \\ 

\cmidrule(lr){2-3}\cmidrule(lr){4-5}\cmidrule(lr){6-7}\cmidrule(lr){8-9}\cmidrule(lr){10-11}\cmidrule(lr){12-13}\cmidrule(lr){14-15}\cmidrule(lr){16-17}\cmidrule(lr){18-19}\cmidrule(lr){20-21}\cmidrule(lr){22-23}
\scalebox{1.2}{Metric} & \scalebox{1.2}{MSE}    & \scalebox{1.2}{MAE}   & \scalebox{1.2}{MSE}    & \scalebox{1.2}{MAE}    & \scalebox{1.2}{MSE}    & \scalebox{1.2}{MAE} & \scalebox{1.2}{MSE}    & \scalebox{1.2}{MAE} & \scalebox{1.2}{MSE}    & \scalebox{1.2}{MAE}  & \scalebox{1.2}{MSE} & \scalebox{1.2}{MAE}      & \scalebox{1.2}{MSE}    & \scalebox{1.2}{MAE}  & \scalebox{1.2}{MSE}    & \scalebox{1.2}{MAE} & \scalebox{1.2}{MSE} & \scalebox{1.2}{MAE}    & \scalebox{1.2}{MSE} & \scalebox{1.2}{MAE}    & \scalebox{1.2}{MSE} & \scalebox{1.2}{MAE}     \\ 
\toprule

\scalebox{1.35}{BE}     & \scalebox{1.35}{\textcolor{red}{\textbf{0.338}}} &  \scalebox{1.35}{\textcolor{blue}{\underline{0.239}}}            
& \scalebox{1.35}{\textcolor{blue}{\underline{0.356}}}    & \scalebox{1.35}{0.250} & \scalebox{1.35}{0.367}    &  \scalebox{1.35}{\textcolor{red}{\textbf{0.231}}} & \scalebox{1.35}{0.608}              & \scalebox{1.35}{0.413} & \scalebox{1.35}{0.411}                 & \scalebox{1.35}{0.288} & \scalebox{1.35}{0.417}          & \scalebox{1.35}{0.269} & \scalebox{1.35}{0.463}              & \scalebox{1.35}{0.333} & \scalebox{1.35}{0.434}             & \scalebox{1.35}{0.336} & \scalebox{1.35}{0.481}                         & \scalebox{1.35}{0.351} & \scalebox{1.35}{0.392}  & \scalebox{1.35}{0.280} & \scalebox{1.35}{0.829} & \scalebox{1.35}{0.549}          \\ \midrule

\scalebox{1.35}{DE}  & \scalebox{1.35}{\textcolor{red}{\textbf{0.266}}}  & \scalebox{1.35}{\textcolor{red}{\textbf{0.322}}} &    \scalebox{1.35}{\textcolor{blue}{\underline{0.268}}} & \scalebox{1.35}{\textcolor{blue}{\underline{0.324}}}        &   
\scalebox{1.35}{0.301} & \scalebox{1.35}{0.328}       &  
\scalebox{1.35}{0.921} & \scalebox{1.35}{0.633}       &          \scalebox{1.35}{0.392} & \scalebox{1.35}{0.399}    & \scalebox{1.35}{0.429} & \scalebox{1.35}{0.416}  & \scalebox{1.35}{0.336} & \scalebox{1.35}{0.351}             & \scalebox{1.35}{0.302} & \scalebox{1.35}{0.323}          & \scalebox{1.35}{0.291} & \scalebox{1.35}{0.333} & \scalebox{1.35}{0.375} & \scalebox{1.35}{0.387}     & \scalebox{1.35}{1.080} & \scalebox{1.35}{0.695}           \\ \midrule

\scalebox{1.35}{FR}     & \scalebox{1.35}{\textcolor{red}{\textbf{0.346}}} & \scalebox{1.35}{\textcolor{red}{\textbf{0.187}}}   
& \scalebox{1.35}{0.371}             & 
\scalebox{1.35}{0.205} & \scalebox{1.35}{\textcolor{blue}{\underline{0.361}}}      &     
 \scalebox{1.35}{\textcolor{blue}{\underline{0.197}}} & \scalebox{1.35}{0.573}                 & \scalebox{1.35}{0.350} & \scalebox{1.35}{0.457}   & \scalebox{1.35}{0.299} & \scalebox{1.35}{0.409}    & \scalebox{1.35}{0.244} & \scalebox{1.35}{0.420}             & \scalebox{1.35}{0.238} & \scalebox{1.35}{0.420}      & \scalebox{1.35}{0.219} & \scalebox{1.35}{0.408} & \scalebox{1.35}{0.236} & \scalebox{1.35}{0.411}    & \scalebox{1.35}{0.273} & \scalebox{1.35}{0.763}  & \scalebox{1.35}{0.480}        \\ \midrule
 
\scalebox{1.35}{NP}    & \scalebox{1.35}{\textcolor{red}{\textbf{0.183}}}  & 
\scalebox{1.35}{\textcolor{red}{\textbf{0.227}}} & \scalebox{1.35}{0.200}     &      
\scalebox{1.35}{0.253} & \scalebox{1.35}{\textcolor{blue}{\underline{0.197}}}  & 
\scalebox{1.35}{\textcolor{blue}{\underline{0.241}}} & \scalebox{1.35}{1.194}    & \scalebox{1.35}{0.799} & \scalebox{1.35}{0.252}   & \scalebox{1.35}{0.287} & \scalebox{1.35}{0.549}     & \scalebox{1.35}{0.554} & \scalebox{1.35}{0.313}   & \scalebox{1.35}{0.367} & \scalebox{1.35}{0.466}              & \scalebox{1.35}{0.478} & \scalebox{1.35}{0.231}   & \scalebox{1.35}{0.296} & \scalebox{1.35}{0.237}   & \scalebox{1.35}{0.273} & \scalebox{1.35}{0.819}  & \scalebox{1.35}{0.657}  \\ \midrule

\scalebox{1.35}{PJM}    & \scalebox{1.35}{\textcolor{red}{\textbf{0.069}}}  
& \scalebox{1.35}{\textcolor{red}{\textbf{0.165}}} & \scalebox{1.35}{\textcolor{blue}{\underline{0.078}}}       &    
\scalebox{1.35}{\textcolor{blue}{\underline{0.166}}} & \scalebox{1.35}{0.082}    &    
\scalebox{1.35}{0.179} & \scalebox{1.35}{0.339}      &     
\scalebox{1.35}{0.323} & \scalebox{1.35}{0.098}   & \scalebox{1.35}{0.215} & \scalebox{1.35}{0.098}     & \scalebox{1.35}{0.193} & \scalebox{1.35}{0.080}   & \scalebox{1.35}{0.176} & \scalebox{1.35}{0.109}   & \scalebox{1.35}{0.167} & \scalebox{1.35}{0.103}   & \scalebox{1.35}{0.177} & \scalebox{1.35}{0.091}   & \scalebox{1.35}{0.206} & \scalebox{1.35}{0.334} & \scalebox{1.35}{0.441}  \\ \midrule

\scalebox{1.35}{{\emph{AVG}}}   & \scalebox{1.35}{\textcolor{red}{\textbf{0.240}}}  & 
\scalebox{1.35}{\textcolor{red}{\textbf{0.228}}} & \scalebox{1.35}{\textcolor{blue}{\underline{0.255}}}  &  
\scalebox{1.35}{0.240} & \scalebox{1.35}{0.261} &          
\scalebox{1.35}{\textcolor{blue}{\underline{0.235}}} & \scalebox{1.35}{0.727}    & \scalebox{1.35}{0.504} & \scalebox{1.35}{0.322}     & \scalebox{1.35}{0.298} & \scalebox{1.35}{0.380}    & \scalebox{1.35}{0.335} & \scalebox{1.35}{0.322}   & \scalebox{1.35}{0.293} & \scalebox{1.35}{0.346}   & \scalebox{1.35}{0.305} & \scalebox{1.35}{0.303}  & \scalebox{1.35}{0.278} & \scalebox{1.35}{0.301}     & \scalebox{1.35}{0.284} & \scalebox{1.35}{0.765}    & \scalebox{1.35}{0.564}       \\ 

\bottomrule
\end{tabular}
\end{scriptsize}
\label{tab:epf-forecast}
\end{table*}

\paragraph{Baselines} We include diverse state-of-the-art forecasting models as our baselines, including multimodal models: TimeLLM \cite{jin2023time}, GPT4TS \cite{zhou2023one}, multivariate models: DUET \cite{qiu2025duet}, CrossLinear \cite{zhou2025crosslinear}, Timer-XL \cite{liu2024timer}, TimeXer \cite{wang2024timexer}, TiDE \cite{das2023long}, iTransformer \cite{liu2023itransformer}, univariate models: DLinear \cite{zeng2023transformers}, PatchTST \cite{nie2022time}, TimesNet \cite{wu2022timesnet}, Autoformer \cite{wu2021autoformer}, regression model: LightGBM \cite{ke2017lightgbm}. All baseline models incorporate both historical and future exogenous series for a fair comparison. For multivariate models, future series are integrated as additional input variables. In the case of univariate models, all sequences are concatenated into a single input. Furthermore, multimodal baselines are provided with full textual information extracted from the data mining process.

\paragraph{Implementation Details} We divide the dataset into normal and abnormal samples according to documented real-world failure records. Following a semi-supervised health monitoring setting, only normal data are used for training and validation to model healthy operational patterns. Detection thresholds are determined on normal test sequences under a maximum false alarm rate of 5\%, and the resulting true positive rate is evaluated on abnormal samples. Anomaly detection is conducted by flagging large forecasting errors on recorded flight sequences, where all exogenous variables, including both historical and future values, are taken from ground-truth records rather than predicted, which is akin to the standard Top-K selection logic in time series anomaly detection.
All experiments are conducted on the takeoff phase, with an input window of 6, a forecasting horizon of 18 time steps, and the exogenous sequence length of 24.

\subsection{Main Results}

As presented in Table \ref{tab:main_result}, Aura consistently achieves state-of-the-art performance across all datasets and evaluation metrics, significantly outperforming competitive baselines. In terms of mean squared error, Aura achieves a substantial relative reduction compared with leading specialized models such as DUET and TimeXer. Furthermore, its superior performance on the True Alert Rate (TAR) highlights its high practical utility and deployment potential for real-world aviation predictive maintenance.

It is noteworthy that, in dynamic thresholding frameworks for anomaly detection, lower prediction errors do not always translate directly into higher alert rates, as observed across several baseline comparisons in our results. Nevertheless, Aura consistently maintains the highest True Alert Rate (TAR) across all scenarios. This performance stems from our integration of rich exogenous information, enabling the model to precisely characterize the aircraft's operational state. By accounting for data fluctuations driven by external factors during normal operation, Aura produces more accurate baseline predictions. Conversely, when mechanical malfunctions occur in the valves, these anomalies deviate from patterns expected under legitimate external events. Consequently, the model fails to fit these irregular data points, resulting in larger residuals that facilitate the identification of abnormal states.

\subsection{Exogenous Forecasting Results}

To evaluate the generality of Aura, we conduct experiments on the EPF benchmark, leveraging both historical and future exogenous series. As listed in Table \ref{tab:epf-forecast}, Aura consistently outperforms state-of-the-art baselines, highlighting its effectiveness in integrating exogenous time series data. These results indicate that the proposed unified framework can flexibly accommodate different configurations of exogenous inputs. Such adaptability is particularly important for real-world forecasting scenarios, where the availability and types of exogenous information can vary, and models are expected to perform reliably even with limited or partial external inputs.

\begin{figure}[h]
  \centering
  \includegraphics[width=1\linewidth]{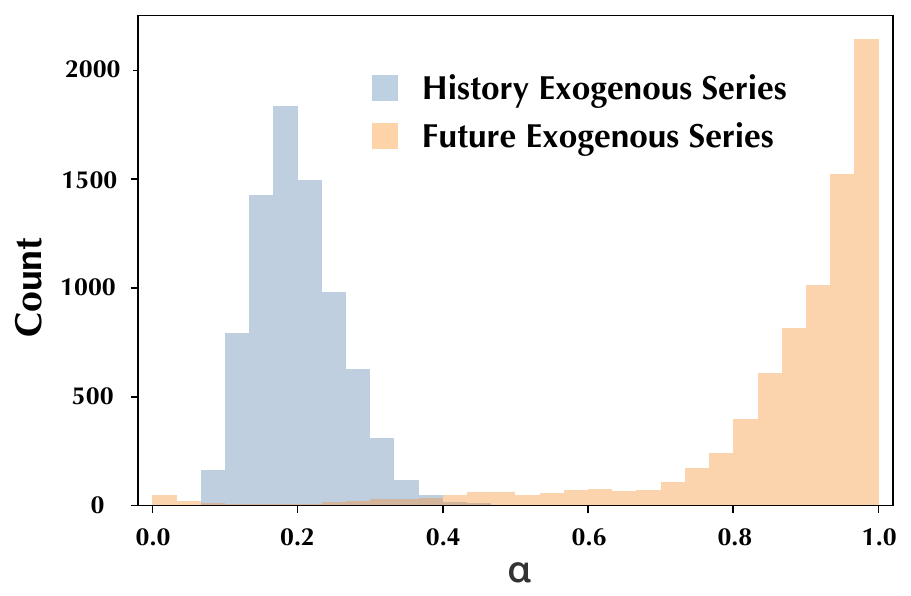}
  \vspace{-15pt}
  \caption{Statistical distribution of gated weights for history and future exogenous series.}
  \label{fig:residual}
\end{figure}

\begin{table*}[t]
  \caption{Ablation Results on the design of Aura. \emph{Data} refers to the removal of three distinct types of exogenous information. \emph{Arch} represents architectural modifications, where the first three experiments employ uniform modeling for all exogenous information, and the final experiment evaluates the removal of the gated residual module.}
  \renewcommand{\arraystretch}{1}
  \centering
  \begin{threeparttable}
  \renewcommand{\multirowsetup}{\centering}
  \setlength{\tabcolsep}{6pt}
  \begin{tabular}{cc|ccc|ccc|ccc|ccc}
    \toprule
    
    \multicolumn{2}{c|}{\multirow{2}{*}{Design}} &
    \multicolumn{3}{c}{\multirow{1}{*}{\scalebox{1.0}{Boeing 777 Left}}} & 
    \multicolumn{3}{c}{\multirow{1}{*}{\scalebox{1.0}{Boeing 777 Right}}} &
    \multicolumn{3}{c}{\multirow{1}{*}{\scalebox{1.0}{Airbus A320 Left}}} &
    \multicolumn{3}{c}{\multirow{1}{*}{\scalebox{1.0}{Airbus A320 Right}}} \\
    
    \cmidrule(lr){3-5} \cmidrule(lr){6-8}\cmidrule(lr){9-11} \cmidrule(lr){12-14}

    & & \scalebox{1.0}{MSE} & \scalebox{1.0}{MAE} & \scalebox{1.0}{TAR} & \scalebox{1.0}{MSE} & \scalebox{1.0}{MAE} & \scalebox{1.0}{TAR} & \scalebox{1.0}{MSE} & \scalebox{1.0}{MAE} & \scalebox{1.0}{TAR} & \scalebox{1.0}{MSE} & \scalebox{1.0}{MAE} & \scalebox{1.0}{TAR} \\
    \toprule

    \multicolumn{2}{c|}{\textbf{Aura}} & \textbf{0.075} & \textbf{0.180} & \textbf{0.625} & \textbf{0.086} & \textbf{0.182} & \textbf{0.625} & \textbf{0.048} & \textbf{0.138} & \textbf{0.345} & \textbf{0.063} & \textbf{0.146} & \textbf{0.414} \\ 
    \midrule
    
    \multicolumn{1}{c}{\multirow{4}{*}{\emph{Data}}} & \multicolumn{1}{c|}{w/o Static Attributes} & 0.107 & 0.213 & 0.500 & 0.094 & 0.200 & \textbf{0.625} & 0.050 & 0.141 & \textbf{0.345} & 0.152 & 0.152 & 0.310 \\ 
    \cmidrule(lr){2-14}
    & \multicolumn{1}{c|}{w/o Dynamic Events} & 0.085 & 0.190 & 0.500 & 0.096 & 0.209 & \textbf{0.625} & 0.050 & 0.139 & \textbf{0.345} & 0.064 & 0.147 & 0.310 \\ 
    \cmidrule(lr){2-14}
    & \multicolumn{1}{c|}{w/o Exogenous Series} & 0.542 & 0.528 & 0.375 & 0.422 & 0.451 & 0.500 & 0.055 & 0.145 & 0.276 & 0.068 & 0.154 & 0.241 \\ 
    \midrule
    \multicolumn{1}{c}{\multirow{5}{*}{\emph{Arch}}} & \multicolumn{1}{c|}{Token Concat} & 0.094 & 0.196 & 0.500 & 0.094 & 0.194 & \textbf{0.625} & 0.054 & 0.142 & 0.276 & 0.064 & 0.149 & 0.310 \\ 
    \cmidrule(lr){2-14}
    & \multicolumn{1}{c|}{Cross Attention} & 0.137 & 0.253 & 0.500 & 0.098 & 0.200 & \textbf{0.625} & 0.066 & 0.154 & 0.310 & 0.064 & 0.148 & 0.310 \\ 
    \cmidrule(lr){2-14}
    & \multicolumn{1}{c|}{Mixture of Experts} & 0.418 & 0.463 & 0.375 & 0.336 & 0.410 & 0.500 & 0.066 & 0.153 & 0.310 & 0.068 & 0.151 & 0.310 \\ 
    \cmidrule(lr){2-14}
    & \multicolumn{1}{c|}{w/o Gated Residual} & 0.085 & 0.192 & \textbf{0.625} & 0.088 & 0.187 & \textbf{0.625} & 0.051 & 0.140 & 0.310 & 0.065 & 0.147 & 0.310 \\ 
    
    \bottomrule
  \end{tabular}
  \end{threeparttable}
\label{tab:ablation_result}
\end{table*}

\subsection{Analysis of the Gating Behavior}

To investigate how the model prioritizes exogenous information across flight phases, we analyze the learned gating coefficients in the residual modules using Boeing 777 Left test dataset. As shown in Figure \ref{fig:residual}, the statistical analysis reveals that the gating weights corresponding to the future series are significantly higher than those assigned to the historical series.

This observation provides strong evidence for the physical interpretability of Aura. Future exogenous series correspond to the climb phase, where the PRSOV system experiences intense aero-thermal loads and rapid thrust transitions. In these stages, future exogenous series, such as engine rotational speed and intermediate pressure, provide crucial predictive cues. Conversely, historical series largely represent the taxiing phase, which typically involves more stable operating conditions. Under these conditions, exogenous factors contribute less to forecasting performance compared to the active flight phases. This alignment between gating behavior and flight operations demonstrates that Aura effectively manifests underlying physical laws through its internal weights, supporting its interpretability in complex industrial scenarios.

\subsection{Ablation Study}

To evaluate the effectiveness of each component in the Aura framework, we conduct a comprehensive ablation analysis as summarized in Table \ref{tab:ablation_result}. We first investigate the necessity of incorporating multi-dimensional data by systematically removing three distinct types of exogenous information (denoted as \emph{Data}). Across all datasets, excluding any exogenous source leads to consistent degradation in forecasting performance, indicating that these external factors provide complementary information for time series modeling.

We further validate architectural design choices by contrasting the proposed multi-aspect exogenous integration with uniform integration strategies (denoted as \emph{Arch}), where all exogenous data are introduced through the same mechanism, including Token Concat, Cross Attention, and Mixture of Experts.
Although the true alert rate may appear relatively stable on datasets with limited anomaly instances, the full Aura model consistently achieves higher forecasting precision and produces more reliable anomaly alerts overall. Additionally, we evaluate the impact of the gated residual module. Results indicate that failing to balance historical and future exogenous series leads to suboptimal performance.

\subsection{Deployed System Evaluation}

To evaluate Aura’s practical efficacy, we deployed it within the operational Aircraft Health Management System of China Southern Airlines, monitoring Boeing 777 and Airbus A320 fleets. During over two months of real-time production, the model maintained high stability with zero false alarms across 13000 flights.

Notably, on December 31, 2025, Aura triggered a critical alert for aircraft B-2XXX, later confirmed as a genuine failure by maintenance records. By enabling proactive intervention, this alert successfully prevented a potential delay or cancellation. According to Boeing’s investigation, one single valid detection would save the airline approximately \$50,000. This substantial cost reduction underscores Aura’s immense commercial value and its transformative potential for large-scale industrial applications.

\begin{figure}[h]
  \centering
  \includegraphics[width=1\linewidth]{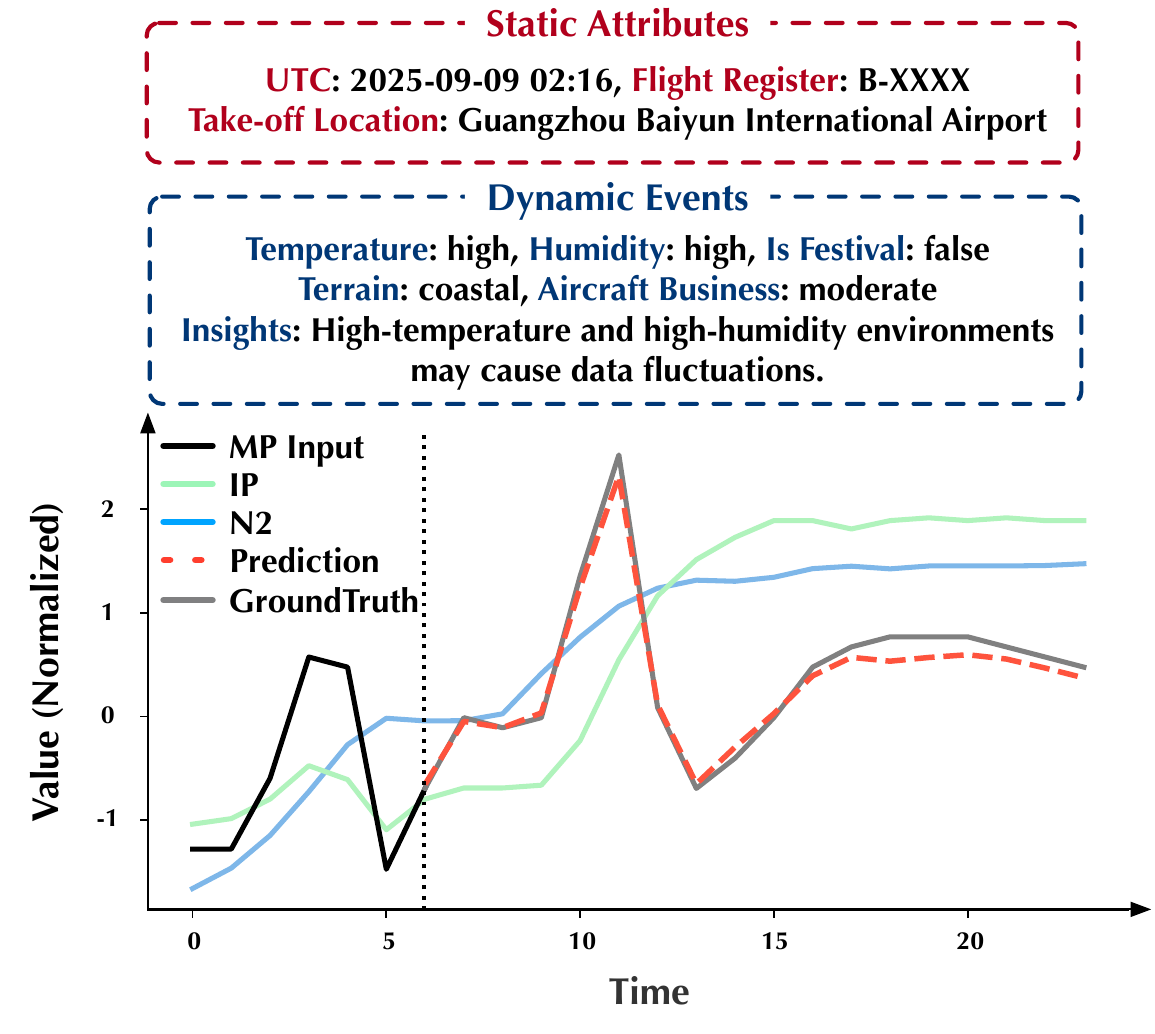}
  \vspace{-15pt}
  \caption{Deployed model inference visualization within the health monitoring system of China Southern Airlines.}
  \label{fig:showcase}
\end{figure}

Figure~\ref{fig:showcase} presents a representative example of system monitoring on a Boeing~777 during the takeoff phase. The predicted values closely align with the observed sensor measurements under the rapid thrust transitions phase.

\section{Conclusion}

In this paper, we introduce Aura, a multi-dimensional exogenous integration framework for time series forecasting and anomaly detection in complex industrial systems. Aura provides a unified framework that systematically incorporates multi-dimensional and multimodal exogenous information through meticulous integration designs, enabling effective modeling of complex industrial time series. Taking the aviation domain as a representative case study, we demonstrate the applicability of Aura to real-world aircraft bleed air monitoring. 
Experimental results on real-world aircraft datasets demonstrate that Aura consistently achieves state-of-the-art performance, providing a robust and reliable solution for health monitoring in practical scenarios.

\bibliographystyle{ACM-Reference-Format}
\bibliography{sample-base}

\appendix
\section{Datasets Description}

\begin{table*}[h]
\caption{Statistics of the datasets used in our experiments.}
\label{tab:dataset_statistics}
\centering
\begin{tabular}{ccccccc}
\toprule
Dataset & Samples & Vars & En. Input & Ex. Input & Output & En. Description \\
\midrule
Boeing 777 Left & (28486, 4181, 7827)  & 3 & 6 & 24 & 18 & Bleed Air Manifold Pressure \\
Boeing 777 Right & (28486, 4181, 7827)  & 3 & 6 & 24 & 18 & Bleed Air Manifold Pressure \\
Airbus A320 Left & (97011, 16401, 30880)  & 3 & 6 & 24 & 18 & Precooler Pressure  \\
Airbus A320 Right & (97011, 16401, 30880)  & 3 & 6 & 24 & 18 & Precooler Pressure  \\
NP   & (36500, 5219, 10460) & 3 & 168 & 192 & 24 & Nord Pool Electricity Price \\
PJM   & (36500, 5219, 10460) & 3 & 168 & 192 & 24 & \makecell{Pennsylvania-New Jersey-Maryland \\ Electricity Price} \\
BE   & (36500, 5219, 10460) & 3 & 168 & 192 & 24 & Electricity Price of Belgium \\
FR   & (36500, 5219, 10460) & 3 & 168 & 192 & 24 & Electricity Price of France \\
DE   & (36500, 5219, 10460) & 3 & 168 & 192 & 24 & Electricity Price of Germany \\
\bottomrule
\end{tabular}
\end{table*}

\subsection{China Southern Airlines Dataset}

We evaluate Aura on a large-scale real-world aviation dataset collected from China Southern Airlines (CSA).
The dataset spans approximately three years of operational records and is constructed for aircraft predictive maintenance and health monitoring tasks.
It covers two major commercial aircraft fleets, including Boeing 777 and Airbus A320, and contains multivariate time series collected under diverse real-world operating conditions.
All measurements are recorded at the flight level, reflecting realistic temporal dynamics and maintenance-related patterns encountered in commercial aviation operations. In our experimental setup, we divided the dataset into training, validation, and test sets according to specific time periods. The training set comprises data collected from January 1, 2023, to March 1, 2025; the validation set consists of data collected from March 1, 2025, to June 1, 2025; and the remaining more recent data are used as the test set. More detailed statistics are provided in Table~\ref{tab:dataset_statistics}. Additionally, it should be noted that both the Boeing 777 datasets and the Airbus A320 dataset describe mechanisms that are related to the same underlying physical principles, as illustrated in Figure~\ref{prsov} in the main text. The inconsistencies in variable descriptions in Table~\ref{tab:dataset_statistics} arise solely from differences in the respective onboard system logging conventions of the two aircraft types.

Given historical multivariate sensor measurements, the forecasting task aims to predict future values of target health-related indicators over a predefined prediction horizon.
For each sample, the model takes a fixed-length historical window as input and outputs the subsequent future segment.
To account for structural and operational differences, we separately consider the left- and right-system configurations for each aircraft type, yielding four datasets: Boeing 777 Left, Boeing 777 Right, Airbus A320 Left, and Airbus A320 Right.

The dataset includes a rich set of endogenous time series variables, such as engine performance and operational sensor measurements, along with auxiliary contextual information.
These data characteristics make the CSA dataset a representative benchmark for evaluating time series forecasting methods in industrial predictive maintenance scenarios.

\subsection{EPF Dataset}

In addition to the aviation dataset, we evaluate Aura on the Electricity Price Forecasting (EPF) dataset, a widely used public benchmark for multivariate time series forecasting with exogenous variables.
The EPF dataset consists of electricity price series sampled at a regular temporal frequency and has been extensively adopted in prior forecasting studies.
To ensure a fair comparison, we follow the official preprocessing procedures and experimental settings used in TimeXer, including the input and prediction-horizon configurations.

\subsection{Dataset Statistics}

Table~\ref{tab:dataset_statistics} summarizes the key statistics of the datasets used in our experiments.
For the CSA dataset, we report the number of samples, the dimensionality of multivariate time series, the input and output lengths, and the description for each aircraft-engine configuration.
The EPF dataset statistics are also included for completeness.

\begin{figure*}[h]
  \centering
  \includegraphics[width=\linewidth]{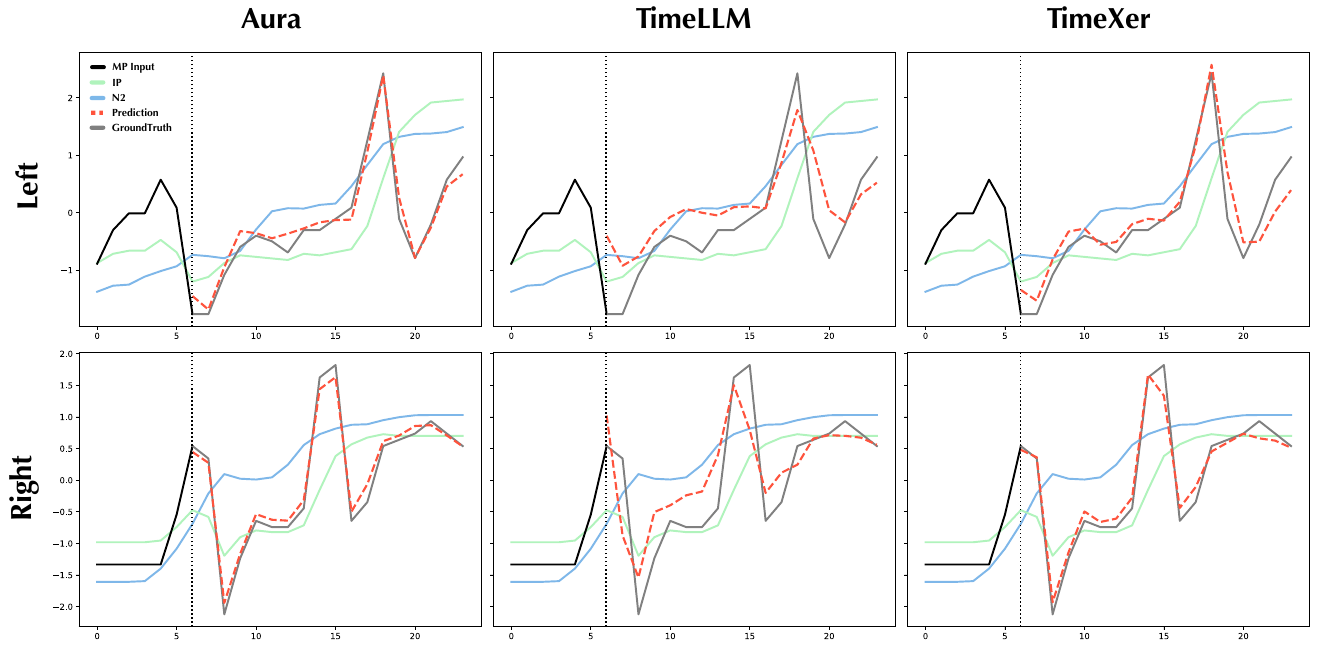}
  \caption{Visualization results of Aura, TimeLLM and TimeXer on Boeing 777 Datasets.}
  \label{fig:vs}
\end{figure*}

\section{Implementation Details}
All the experiments are implemented in PyTorch \cite{paszke2019pytorch} and conducted on a single NVIDIA 4090 24GB GPU. We utilize ADAM \cite{kingma2014adam} with an initial learning rate $5\times 10^{-4}$ for CSA datasets and $10^{-4}$ for EPF datasets and L2 loss for the model optimization. The training process is fixed to 100 epochs for CSA and 10 epochs for EPF with early stopping. We set the number of the transformers layers in our proposed model $L \in \{1, 2, 3\}$. The dimension of series representations $d_{model}$ is searched from $\{128, 256, 512\}$. The patch length is uniformly set to 6 for CSA and 24 for EPF. The batch size in our experiments is selected from $\{420, 840\}$ for CSA and from $\{4, 16\}$ for EPF. We reproduced the compared baseline models based on the benchmark of TimeXer \cite{wang2024timexer} Repository.

We employ BERT \cite{devlin2019bert} to encode static attributes, while Qwen2.5-0.5B-Instruct \cite{qwen2025qwen25technicalreport} is utilized to perform inference on dynamic contextual events.

To evaluate the performance of our model, we utilize two widely recognized metrics in the field of predictive modeling: Mean Squared Error (MSE) and Mean Absolute Error (MAE). For a given time series $\mathbf{x}_{1:L} = \{\mathbf{x}_1, \mathbf{x}_2, \dots, \mathbf{x}_L\}$, where $\widehat{\mathbf{x}}_{i}$ denotes the corresponding predicted value at time step $i$, these metrics are formally expressed as:
\begin{equation}\label{equ:metrics}
\begin{aligned}
\text{MSE} &= \sum_{i=1}^L |\mathbf{x}_{i} - \widehat{\mathbf{x}}_{i}|^2, \\
\text{MAE} &= \sum_{i=1}^L |\mathbf{x}_{i} - \widehat{\mathbf{x}}_{i}|.
\end{aligned}
\end{equation}

\section{Details of Textual Information Process}

\subsection{Context Sources and Scope}

Instead of using free-form maintenance narratives, Aura leverages a limited set of structured external operational context signals.
These signals describe high-level operating conditions that may influence system-level regulation behavior, rather than component faults or maintenance actions.
Specifically, we consider four contextual factors: operational load, ambient temperature, humidity, and terrain characteristics.

This design intentionally restricts the scope of textual information to non-diagnostic and non-fault-related context, ensuring that the textual input serves only as auxiliary environmental information complementary to the numerical time series.

\subsection{Context Extraction and Discretization}

Raw contextual inputs are obtained from timestamps, environmental measurements, and airport-related metadata when available.
Temporal information is first standardized to Coordinated Universal Time (UTC), from which weekend indicators are derived.
Public holiday information is obtained either from external metadata or an external calendar library when available; otherwise, it is marked as unknown.

Continuous environmental variables, including ambient temperature and humidity, are discretized into coarse categorical levels using predefined rules.
Terrain characteristics are determined based on explicit terrain type labels when available, or approximated using airport elevation metadata.
If any contextual attribute is missing, it is explicitly labeled as unknown to avoid introducing spurious assumptions.

This rule-based discretization step transforms heterogeneous raw inputs into a small set of interpretable categorical context variables, reducing noise and improving robustness.

\subsection{Prompt-Guided Contextual Summarization}
\begin{figure*}[h]
  \centering
  \includegraphics[width=\linewidth]{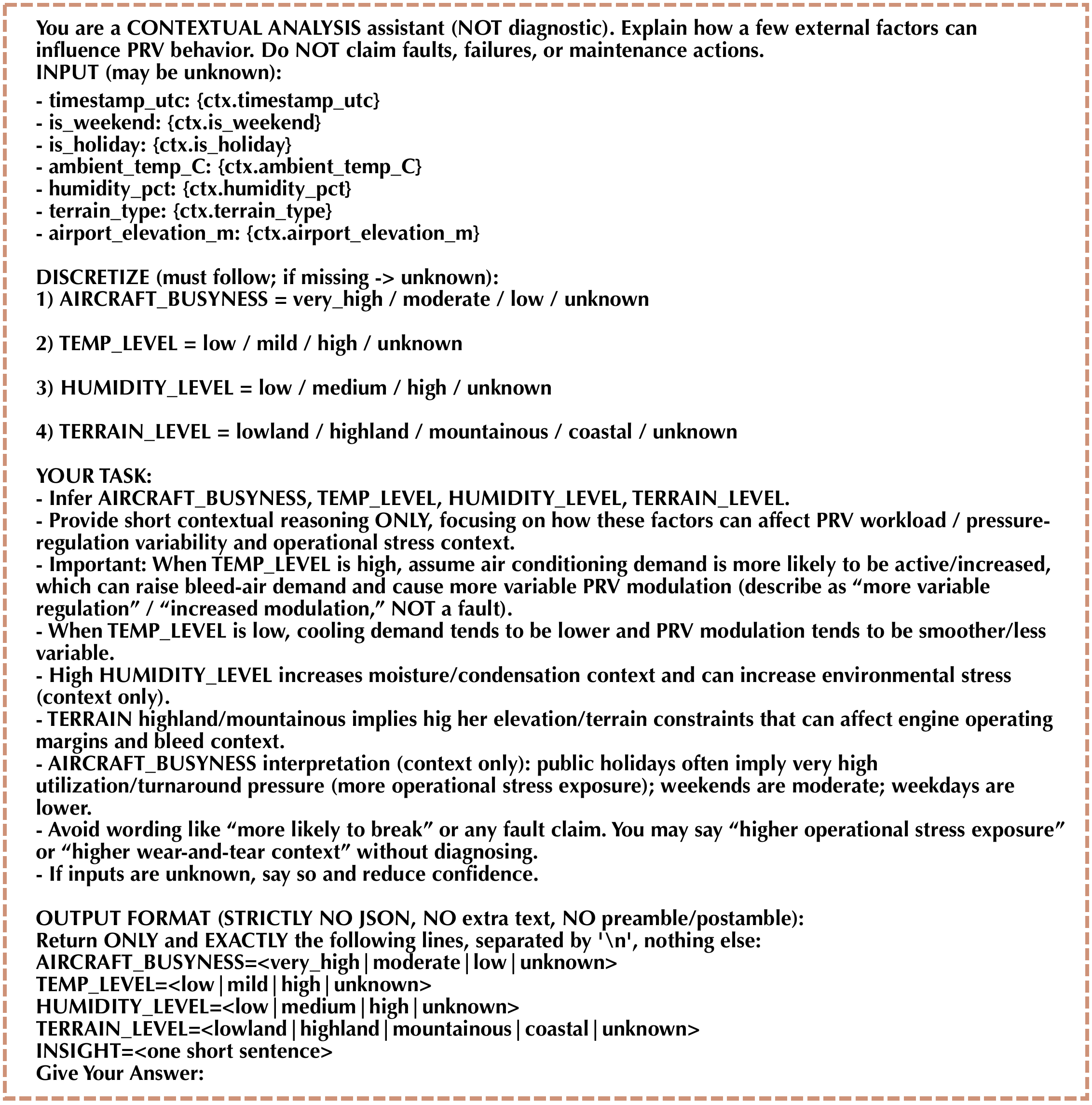}
  \caption{The Prompt Design of Aura for dynamic contextual events.}
  \label{fig:prompt}
\end{figure*}

Figure~\ref{fig:prompt} presents the prompt design for the LLM in Aura, which facilitates more effective utilization of authentic information.
The discretized contextual variables are converted into a structured English prompt and fed into a frozen large language model (LLM).
The prompt explicitly defines the allowable interpretations and constraints, instructing the LLM to provide only contextual explanations of how external operating conditions may influence pressure regulation behavior.
Importantly, the prompt prohibits any diagnostic statements, fault attribution, or maintenance recommendations.

The LLM output consists of a fixed set of categorical context labels and a single concise explanatory sentence.
This constrained output format ensures low variability and high reproducibility across samples, while still allowing the model to generate semantically meaningful summaries of operational context.

\begin{table}[t]
  \caption{Error Bar Analysis of Aura. The results are obtained from three random runs.}
  \label{tab:robustness}
  \centering
  \begin{threeparttable}
    \footnotesize 
    \setlength{\tabcolsep}{0pt} 
    \begin{tabular*}{\columnwidth}{@{\extracolsep{\fill}} cccc @{}}
      \toprule
      \textbf{Data}  & \textbf{MSE} & \textbf{MAE} & \textbf{TAR} \\
      \midrule
      {Boeing 777 Left} 
        & $0.075 \pm 0.004$ & $0.180 \pm 0.004$ & $0.625 \pm 0.000$ \\
        \midrule
        {Boeing 777 Right} 
       & $0.086 \pm 0.005$ & $0.182 \pm 0.010$ & $0.625 \pm 0.000$ \\
      \midrule
      {Airbus A320 Left} 
        & $0.048 \pm 0.001$ & $0.138 \pm 0.000$ & $0.345 \pm 0.000$ \\
        \midrule
        {Airbus A320 Right} 
       & $0.063 \pm 0.004$ & $0.146 \pm 0.002$ & $0.414 \pm 0.000$ \\
      \bottomrule
    \end{tabular*}
  \end{threeparttable}
\end{table}

\section{Robustness Analysis}

To evaluate Aura's robustness, we run experiments across multiple random seeds.
All models are trained and evaluated under identical settings, and the mean and standard deviation of forecasting performance are reported.
This analysis assesses the stability of Aura under random initialization and stochastic optimization.

Table~\ref{tab:robustness} reports the results of Aura in three different random seeds. The consistently low variance indicates that Aura yields stable performance gains.

\section{Showcase}

We present qualitative case studies to demonstrate Aura's effectiveness in real-world forecasting scenarios.
Figure~\ref{fig:showcase} in the main text visualizes representative forecasting results on the CSA dataset.
To more clearly demonstrate the differences among various models, we visualize the prediction results on the Boeing 777 Datasets in Figure~\ref{fig:vs}. We make a fair comparison among three models, Aura, Time LLM, and TimeXer. Among them, Aura achieves the best visual prediction effect on the two datasets.

\begin{figure}[h]
  \centering
  \includegraphics[width=1\linewidth]{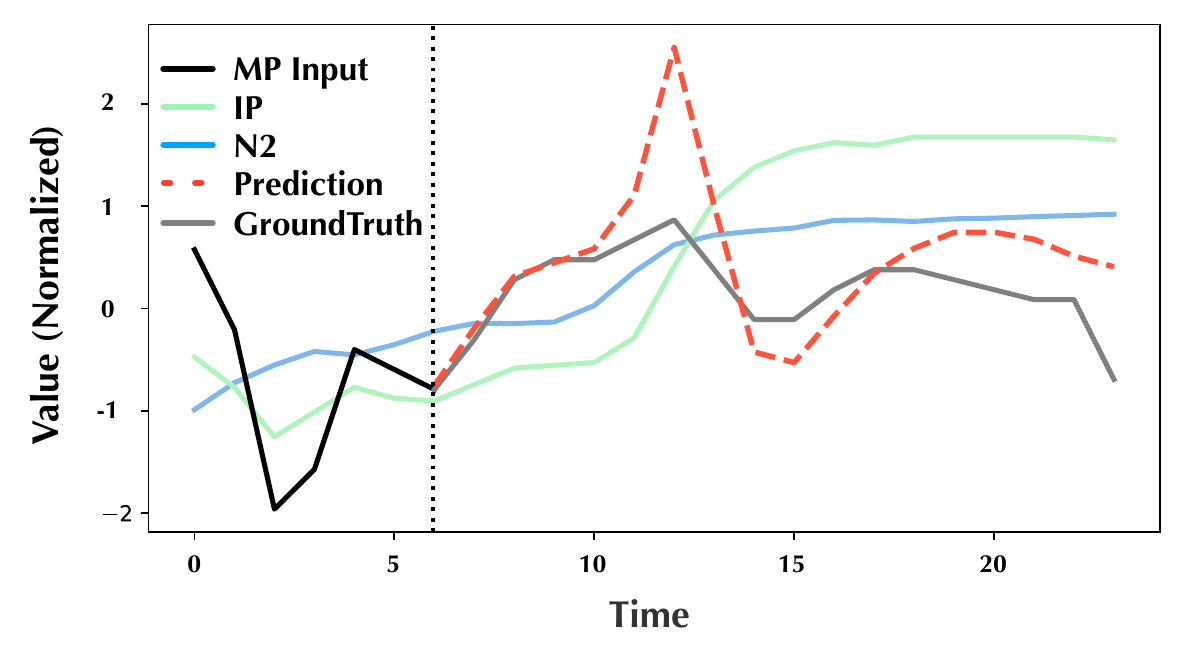}
  \vspace{-15pt}
  \caption{Visualization Prior to PRSOV Anomaly.}
  \label{fig:failure-showcase}
\end{figure}

Furthermore, we investigate the sensitivity of the model's predictive performance as failure approaches. On October 21, 2024, aircraft B-2XXX experienced a PRSOV malfunction. To evaluate the response of Aura, we analyze flight data from October 19, 2024, which is two days prior to the incident. As illustrated in Figure \ref{fig:failure-showcase}, a significant divergence emerged between the actual manifold pressure and the predicted values. This pronounces residual indicates that the model can effectively capture subtle degradation signatures well before functional failure. Such predictive accuracy demonstrates substantial potential for proactive maintenance in real-world industrial deployments.

\end{document}